\documentclass{ceurart}
\usepackage{tikz}
\usetikzlibrary{arrows.meta,positioning,shapes.geometric,fit,calc,backgrounds}
\usepackage{listings}
\lstdefinelanguage{ttql}{
  morecomment=[l]\#,
  morestring=[b]",
  sensitive=true,
  alsoletter={-:@?},
  morekeywords={@prefix,pattern,instance-count,sample-size,@literal},
  morekeywords=[2]{prod:Hardware,prod:ProductCategory,prod:Employee,
                   prod:Department,prod:Manager,prod:Supplier,prod:Price,
                   prod:BomPart,prod:BillOfMaterial,prod:Service,prod:Agent},
  morekeywords=[3]{prod:compatibleProduct,prod:hasCategory,
                   prod:hasProductManager,prod:hasManager,prod:hasSupplier,
                   prod:memberOf,prod:name,prod:email,prod:phone,
                   prod:areaOfExpertise,prod:responsibleFor,prod:id,
                   prod:addressText,prod:addressCountry,
                   prod:addressCountryCode,prod:addressLocality,
                   prod:country,rdfs:label}
}
\graphicspath{{./}}

\definecolor{qrakenline}{HTML}{0E7490}
\definecolor{qrakenfill}{HTML}{D6EEF5}
\definecolor{ttqlline}{HTML}{B45309}
\definecolor{ttqlfill}{HTML}{FCE6C0}
\definecolor{ttqlfilllight}{HTML}{FDF3E1}

\begin{document}

\copyrightyear{2026}
\copyrightclause{Copyright for this paper by its authors.
  Use permitted under Creative Commons License Attribution 4.0 International (CC BY 4.0).}

\conference{Graph-enhanced LLMs for trustwOrthy Web data management (GLOW),
  in ISWC 2026 Workshops Joint Proceedings, October 25--26, 2026, Bari, Italy}

\title{Natural Language Questions as an Interface for Knowledge Graphs:
QRAKEN Graph Distillation and Semantic Self-Healing}

\author[1,2]{Remo Grillo}[%
  orcid=0000-0003-3585-0198,
  email=remo.grillo@unibo.it,
  url=https://remogrillo.me,
]
\cormark[1]

\author[2]{Lukas Klic}[%
  orcid=0000-0002-9620-7107,
  email=lklic@itatti.harvard.edu,
]

\author[1,3]{Giovanni Colavizza}[%
  orcid=0000-0002-9806-084X,
  email=giovanni.colavizza@unibo.it,
]

\address[1]{University of Bologna, Via Zamboni 33, 40126 Bologna, Italy}
\address[2]{I Tatti --- The Harvard University Center for Italian
  Renaissance Studies, Via di Vincigliata 26, 50135 Florence, Italy}
\address[3]{Department of Communication, University of Copenhagen,
  Karen Blixens Plads 8, 2300 Copenhagen S, Denmark}

\cortext[1]{Corresponding author.}

\begin{abstract}
Natural-language access to RDF knowledge graphs is a core Semantic Web
ambition, and large language models (LLMs) have recently made
Text-to-SPARQL feasible at scale; yet the dominant failure mode, on
graphs outside the LLM pretraining distribution, is not syntax but
\emph{graph-grounded correctness}: queries that parse, use real
vocabulary, and still silently misrepresent the data model of the
target graph. Existing approaches --- online agentic exploration, fine-tuning or
retrieval over question-query traces, and shape/metadata prompting ---
share a common limitation: generation is grounded in what the graph
\emph{should} contain according to an ontology or shape specification,
rather than in how it is actually populated.
We present QRAKEN, a training-free, ontology-agnostic neurosymbolic
pipeline for question answering over RDF knowledge graphs. An offline
distiller summarizes the populated graph
into a TTQL file --- a compact description of empirical
multi-hop patterns, conditional frequencies, and path-conditioned literal
examples --- plus a co-occurrence matrix of classes and properties. Online, the NTQ
module injects TTQL into the LLM context and passes the candidate SPARQL
through deterministic syntactic, lexical, and data-model checks
(grounded purely in the distilled empirical evidence), whose diagnostics
feed a deterministic self-healing refinement loop.
On CK25 (First International Text2SPARQL Challenge), under
matched-condition recomputation on a QLever snapshot, QRAKEN
exhibits a better performance among systems built on the same base
model family: \textbf{strict F1 $0.643\pm0.026$ with GPT-4.1 mini and
$0.652\pm0.012$ with GPT-5.4}, \textbf{+30\%} and \textbf{+32\%} over
the strongest recomputed participant. An ablation isolates
TTQL pattern blocks as the dominant driver ($+0.31$ strict F1
over a shape-only baseline), while the self-healing loop acts as a
cheap deterministic safety net that rejects queries whose triple
patterns have zero support in the co-occurrence matrix.
Against an auto-derived SHACL baseline, TTQL
delivers \textbf{$+64$\%} higher strict F1, confirming that its value
lies in pattern-level empirical content, not mere schema exposure.
The same pipeline driving two 35B 4-bit open-weight models locally, at
zero marginal cost, performs on par with the strongest recomputed
participant system, and both TTQL effects persist.
Results come from a single, relatively small benchmark and should be
read as an initial empirical signal; the main current limitation is
monolithic TTQL injection on very open cross-domain graphs.

\end{abstract}


\maketitle

\section{Introduction}
\label{sec:intro}

Natural-language access to RDF knowledge graphs (KGs) is a long-standing
goal of the Semantic Web and, concretely, of knowledge graph question
answering (KGQA): translating a user's question into SPARQL lets
non-expert users exploit the reasoning and integration guarantees of
RDF-based data without having to master its query language. With the
accelerated adoption of large language models (LLMs)~\cite{khorashadizadeh2024},
Text-to-SPARQL has become the dominant incarnation of this goal, and
recent initiatives such as the First International Text2SPARQL
Challenge~\cite{text2sparql2025}
have turned it into a measurable community task. Yet bringing LLM-based
query generation to the quality level required by real Semantic Web
deployments --- cultural-heritage, enterprise, scientific, and private
graphs that are absent from pretraining corpora and whose populated data
diverges from their declared ontologies --- remains an open research
problem, distinct from general code generation and from Text-to-SQL
precisely because of RDF's open-world semantics and the gap between
prescriptive schemas (RDFS/OWL, SHACL/ShEx) and empirical graph usage.
In realistic deployments, LLMs routinely produce queries that are
syntactically plausible but semantically unfaithful to the target
graph: they hallucinate predicates, invert edge directions, collapse
multi-hop relations into nonexistent shortcuts, or pick projections
that do not match the expected answer form. The dominant difficulty
is therefore not language understanding but
\textbf{graph-grounded correctness}.

In this setting, bare prompting leaves the model under-grounded ---
the target vocabulary and its populated usage are not in the pretraining
distribution --- while fine-tuning requires ontology-specific
question-query pairs that are rarely available for private or evolving
graphs and does not adapt to schema drift.

QRAKEN takes a different route. Instead of teaching the model a graph
through machine learning, it \textbf{summarizes the graph into a compact,
LLM-readable intermediate representation}, then couples generation with
in-context learning and \textbf{deterministic correctness checks}. The
pipeline is \emph{training-free} in the sense adopted throughout the
paper: no question-query training corpus is needed for a new graph,
and the LLM does not need to study the graph at inference time;
graph-side context (the TTQL file) is prepared once offline, and the
online path is a single supervised prompt plus deterministic
validation.

The paper makes three contributions:
\begin{enumerate}
  \item a \textbf{conceptual account} of correctness for Text-to-SPARQL,
    distinguishing syntactic, lexical, and data-model layers and arguing
    that the last is where most LLM failures occur
    (Sect.~\ref{sec:correctness});
  \item \textbf{QRAKEN} as a training-free, graph-distillation-based pipeline
    built around \textbf{TTQL}, a Turtle-inspired descriptive language
    for empirical graph patterns, together with a correctness-guided
    self-healing loop (Sect.~\ref{sec:method});
  \item an \textbf{empirical evaluation} on CK25 showing that, under
    matched-condition recomputation, QRAKEN outperforms the participant
    systems of the First Text2SPARQL Challenge on strict F1
    on this single benchmark, suggesting the potential of the approach
    (Sect.~\ref{sec:eval}).
\end{enumerate}

The broader systems-level claim is that, in domain-specific KGQA,
\emph{graph grounding} matters more than model scale: the results come
not from larger LLMs or elaborate agent choreography, but from giving a
compact model a faithful representation of how the target graph is
actually used --- an indication that warrants validation on additional benchmarks.

\section{Related Work}
\label{sec:related}

The First International Text2SPARQL Challenge standardized only the
external protocol (a REST endpoint returning a SPARQL query within a
fixed timeout), leaving model choice, retrieval, indexing, and repair to
participants. Its submissions and several post-challenge systems span
three design families (Sects.~\ref{sec:agentic}--\ref{sec:shape}); a
fourth dimension, cost (Sect.~\ref{sec:cost}), is largely missing from
current evaluations.

\subsection{Online agentic graph exploration}
\label{sec:agentic}

A first family delegates graph understanding to per-question online
exploration. \textbf{ARUQULA} (INFAI)~\cite{aruqula2025} generalizes a
SPINACH-style ReAct workflow to arbitrary RDF graphs, driving utilities
for searching, inspecting, and querying a hybrid QLever/Qdrant/Lucene
backend; it reports GPT-4.1 mini as best cost-quality trade-off, which
makes it the closest same-model comparator for QRAKEN, but pays a
characteristic $\sim$1\,min/question for iterative tool use.
\textbf{Graf von Data (GvD)}~\cite{gvd2025} adopts a compact
three-action design on open-weight backends (IIS-Q with Qwen 2.5 72B,
IIS-L with Llama 3.3 70B), noting that sub-10B models are insufficient
for reliable agent loops. \textbf{SPINACH}~\cite{spinach} and the
post-challenge \textbf{GRASP}~\cite{grasp} imitate how experts
iteratively write SPARQL on Wikidata, with strong fine-tuning-free
results. Relative to QRAKEN, this family sits at the
\emph{online-exploration} end of the design space, re-deriving graph
understanding at every question. A loosely related Text-to-SQL line (Spider~\cite{spider}) generalizes
to unseen databases, but RDF adds open-world semantics and the
prescriptive/descriptive gap of Sect.~\ref{sec:gap}.

\subsection{Memory- and training-based approaches}
\label{sec:memory}

A second family front-loads effort into stored procedural knowledge or
model weights. \textbf{mKGQAgent} (WSE)~\cite{perevalov2025}, the
overall challenge winner, combines LLM reasoning, entity linking, and
an offline experience pool of successful solution traces retrieved
online; it is also one of the few submissions to report cost
explicitly ($\sim$13 LLM calls/query, \$0.48/100q with GPT-3.5 and
\$3.06 with GPT-4o). Fine-tuned autoregressive models form the
supervised counterpart: the \textbf{DBpedia Group}~\cite{soru2025}
fine-tuned CodeGen-350M/StarCoder-1B/CodeLlama-7B on a curated NSpM
subset, and \textbf{FIRESPARQL}~\cite{firesparql} combines fine-tuned LLMs
with RAG context and a SPARQL-correction layer.
Such adaptation is powerful when annotated data is abundant but less
attractive for private or evolving graphs; D'Abramo et
al.~\cite{dabramo2025} show that in-context learning with retrieved
demonstrations alone can match fine-tuned baselines, reinforcing the
training-free case on which QRAKEN is built. Gashkov et al.~\cite{gashkov2025} further show that SPARQL-generation
accuracy depends on whether benchmark entities appeared in pretraining,
a caveat for DBpedia/Wikidata baselines. More broadly,
KAPING~\cite{kaping}, StructGPT~\cite{structgpt} and PGMR~\cite{pgmr2025}
show that per-question context injection over structured data, and
decoupling structure generation from URI grounding, beat parameter
scaling; QRAKEN follows the same intuition but replaces per-question
retrieval with a one-off graph-level distillation
(Sect.~\ref{sec:distill}) and grounds correctness against the distilled
co-occurrence matrix rather than at retrieval time.

\subsection{Shape- and metadata-informed prompting}
\label{sec:shape}

A third family injects structured graph context into the prompt
without online exploration or training. \textbf{AIFB}~\cite{aifb2025}
shows that supplying extracted data shapes substantially improves
performance on unseen graphs: across three models, non-shape-informed
prompts failed to produce valid SPARQL on CK25 whereas shape-informed
ones did, establishing that graph-aware structural context can matter
more than model memorization.
\textbf{SPARQL-LLM}~\cite{sparqlllm2025}, post-challenge, generalizes
this with a triplestore-agnostic, metadata-driven architecture that
retrieves schema metadata and example queries at query time, generates
SPARQL, and repairs candidates in a loop, reporting higher F1 than the
challenge winners while treating runtime and cost as first-class
dimensions ($\sim$\$0.01/q). \textbf{AIRI}~\cite{airi2025}, a modular
multilingual pipeline, belongs to the same family on its Corporate
branch but is exposed to early-stage error propagation.
A closer precursor is Mountantonakis and Tzitzikas's ontology path
patterns for CIDOC-CRM~\cite{mountantonakis2025}, which also inject
multi-hop structures into prompts; theirs, however, are ontology-derived
and prescriptive, whereas TTQL records what the populated data
contains.
QRAKEN shares this family's intuition --- graph-aware context
injection is essential --- but injects a different \emph{kind} of
context: empirical multi-hop usage patterns, literal examples, and
co-occurrence statistics distilled from the populated graph, rather
than prescriptive shapes or raw metadata.

\subsection{Cost and the open region in the design space}
\label{sec:cost}
\label{sec:gap}

Headline F1 scores hide the cost of obtaining them: systems differ in
model scale, LLM and endpoint calls per query, indexing and
training-data requirements, and latency, yet Text2SPARQL 2025 left
backend choice and cost reporting unconstrained. The three families trace three cost
profiles --- agentic exploration pays at every question, memory and
training amortize into one-off preparation with annotated data, and
shape/metadata prompting pays retrieval overhead --- and each imposes a
structural limitation: re-deriving graph understanding at every query,
needing traces or fine-tuning data rarely available for private graphs,
or injecting \emph{prescriptive} structure rather than
\emph{descriptive} content. Following SPARQL-LLM~\cite{sparqlllm2025}, we also
argue that future evaluations should report LLM calls, tokens,
endpoint calls, latency, and separately-accounted preparation cost
alongside F1, a convention we adopt in Sect.~\ref{sec:eval}. The
region that remains under-explored is therefore a system that is at
once \textbf{training-free}, \textbf{cheap at query time}, and
\textbf{grounded in empirical graph usage}; the rest of the paper
develops one answer, by first formalizing correctness
(Sect.~\ref{sec:correctness}), describing the architecture
(Sect.~\ref{sec:method}), and evaluating on CK25
(Sect.~\ref{sec:eval}).

\section{What Does It Mean for a Query to Be Correct?}
\label{sec:correctness}

We treat correctness in KGQA as a \textbf{layered} notion of
increasing stringency.
\textbf{(1) Syntactic} --- the candidate conforms to the SPARQL 1.1
grammar, with declared prefixes and well-formed clauses; straightforward
for modern LLMs and checkable by any parser, but guarantees only that
the string is \emph{executable in principle}.
\textbf{(2) Lexical / schema-level} --- classes, properties, and
namespaces actually exist in the populated graph. Hallucinated
predicates, classes used in the wrong role, or prefixes borrowed from
more familiar graphs fail here; the failures are insidious because
they often yield \emph{empty} result sets rather than explicit
errors.
\textbf{(3) Data-model} --- semantic correctness against the
\emph{actual} populated data. A model may connect classes through a
property that is real but impossible in that direction, or compress a
multi-hop relation into a nonexistent shortcut. Such queries
\emph{execute and return nothing}, so validation cannot be reduced to
parser checks or execution outcomes.

\begin{sloppypar}
As a concrete CK25 illustration, a question asking for the product
manager of a given hardware category yields the fluent-but-unfaithful
triple \texttt{?category\ a\ prod:ProductCategory\ ;} \linebreak[1]
\texttt{prod:hasProductManager\ ?m\ .}, which parses and uses only real
vocabulary (layers 1--2 pass) but violates the populated data model:
the relation flows through \texttt{Hardware}
(\texttt{ProductCategory}\,$\leftarrow$\,\linebreak[1]\texttt{hasCategory}\,$\leftarrow$\,\linebreak[1]\texttt{Hardware}\,$\rightarrow$\,\linebreak[1]\texttt{hasProductManager}\,$\rightarrow$\,\linebreak[1]\texttt{Employee}),
and a data-model check against the co-occurrence evidence rejects the
\texttt{(ProductCategory,\ \linebreak[1]hasProductManager)} shortcut
because it has zero support in the distilled summary. The core
conceptual shift is therefore to treat LLM output not as authoritative
but as a \emph{candidate hypothesis} that must survive transparent,
deterministic verification against all three layers. Sect.~\ref{sec:ntq}
turns this into an explicit pipeline of operational checks.
\end{sloppypar}

\section{QRAKEN: Method and Architecture}
\label{sec:method}

\subsection{Design overview}
\label{sec:design}

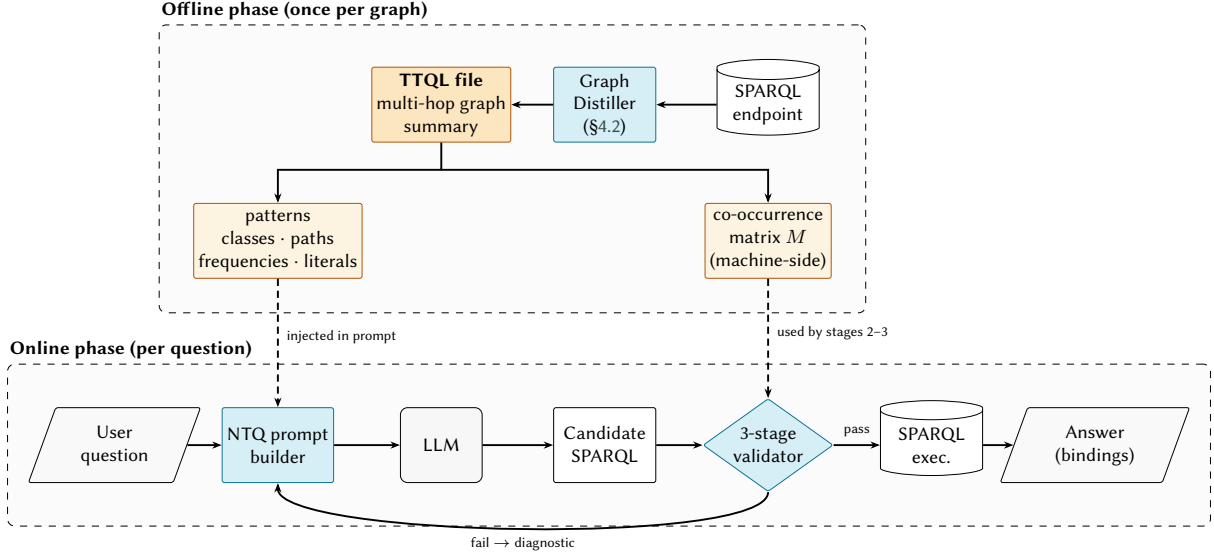
\begin{figure}[t]
  \centering
  \resizebox{\columnwidth}{!}{%
  \begin{tikzpicture}[
      font=\scriptsize,
      >={Stealth[length=4pt]},
      box/.style   ={draw, rounded corners=1pt, align=center,
                     inner sep=2pt,
                     minimum height=\bh, minimum width=15mm,
                     fill=white},
      store/.style ={box, shape=cylinder, shape border rotate=90,
                     aspect=0.25, minimum width=15mm},
      data/.style  ={box, fill=gray!8, minimum width=18mm},
      io/.style    ={box, trapezium, trapezium left angle=70,
                     trapezium right angle=110, fill=gray!5,
                     minimum width=14mm},
      valid/.style ={box, diamond, aspect=1.4, inner sep=1pt, fill=gray!10,
                     minimum width=14mm},
      llm/.style   ={box, rounded corners=3pt, fill=gray!6,
                     minimum width=12mm},
      qraken/.style ={box, fill=qrakenfill, draw=qrakenline},
      qvalid/.style ={valid, fill=qrakenfill, draw=qrakenline},
      ttql/.style   ={data, fill=ttqlfill, draw=ttqlline},
      ttqlpart/.style={data, fill=ttqlfilllight, draw=ttqlline},
      phase/.style ={draw, dashed, rounded corners=3pt,
                     inner sep=4mm, fill=gray!3},
      flow/.style  ={->, thick},
      dflow/.style ={->, thick, densely dashed},
    ]
    %
    \pgfmathsetlengthmacro{\cw}{24mm}   
    \pgfmathsetlengthmacro{\rg}{20mm}   
    \pgfmathsetlengthmacro{\bh}{11mm}   
    \pgfmathsetlengthmacro{\phasegap}{10mm}
    \def\colx#1{#1*\cw}

    \node[io,    at={(\colx{1}, 0)}]   (Q)    {User\\question};
    \node[qraken, at={(\colx{2}, 0)}]  (NTQ)  {NTQ prompt\\builder};
    \node[llm,   at={(\colx{3}, 0)}]   (LLM)  {LLM};
    \node[box,   at={(\colx{4}, 0)}]   (CAND) {Candidate\\SPARQL};
    \node[qvalid, at={(\colx{5}, 0)}]  (V)    {3-stage\\validator};
    \node[store, at={(\colx{6}, 0)}]   (EXEC) {SPARQL\\exec.};
    \node[io,    at={(\colx{7}, 0)}]   (ANS)  {Answer\\(bindings)};

    \draw[flow] (Q)    -- (NTQ);
    \draw[flow] (NTQ)  -- (LLM);
    \draw[flow] (LLM)  -- (CAND);
    \draw[flow] (CAND) -- (V);
    \draw[flow] (V)    -- node[above,font=\tiny]{pass} (EXEC);
    \draw[flow] (EXEC) -- (ANS);

    \draw[flow] (V.south) .. controls +(0,-8mm) and +(0,-8mm) ..
                node[pos=0.5,below,font=\tiny,align=center]
                {fail $\rightarrow$ diagnostic}
                (NTQ.south);

    \begin{scope}[on background layer]
      \node[phase, fit=(Q)(NTQ)(LLM)(CAND)(V)(EXEC)(ANS),
            inner sep=5mm,
            label={[anchor=south west,font=\scriptsize\bfseries,
                    inner sep=1pt,align=left,yshift=1pt]
                    north west:Online phase (per question)}]
            (ON) {};
    \end{scope}

    \node[ttqlpart, at={(\colx{2}, \rg+\phasegap)}] (PAT)
         {patterns\\\scriptsize classes $\cdot$ paths\\\scriptsize frequencies $\cdot$ literals};
    \node[ttqlpart, at={(\colx{5}, \rg+\phasegap)}] (M)
         {co-occurrence\\matrix $M$\\\scriptsize(machine-side)};

    \node[ttql,  at={(\colx{3}, 2*\rg+\phasegap)}] (TTQL)
         {\textbf{TTQL file}\\\scriptsize multi-hop graph\\\scriptsize summary};
    \node[qraken, at={(\colx{4}, 2*\rg+\phasegap)}] (D)
         {Graph\\Distiller\\(\S\ref{sec:distill})};
    \node[store, at={(\colx{5}, 2*\rg+\phasegap)}] (EP)
         {SPARQL\\endpoint};

    \draw[flow] (EP) -- (D);
    \draw[flow] (D)  -- (TTQL);

    \coordinate (FORK) at ($ (TTQL.south) + (0,-4mm) $);
    \draw[flow] (TTQL.south) -- (FORK) -| (PAT.north);
    \draw[flow] (FORK)       -| (M.north);

    \begin{scope}[on background layer]
      \node[phase, fit=(EP)(D)(TTQL)(PAT)(M),
            inner sep=5mm,
            label={[anchor=south west,font=\scriptsize\bfseries,
                    inner sep=1pt,align=left,yshift=1pt]
                    north west:Offline phase (once per graph)}]
            (OFF) {};
    \end{scope}

    \draw[dflow] (PAT.south) --
                 node[right,font=\tiny,pos=0.45]{injected in prompt}
                 (NTQ.north);
    \draw[dflow] (M.south) --
                 node[right,font=\tiny,pos=0.45]{used by stages 2--3}
                 (V.north);
  \end{tikzpicture}%
  }
  \caption{QRAKEN architecture. The \emph{offline phase} (top) distills
  the populated graph into a TTQL file (injected into the LLM prompt)
  and a co-occurrence matrix (kept machine-side for deterministic
  checks). The \emph{online phase} (bottom) couples LLM-based SPARQL
  generation with a three-stage correctness-guided self-healing loop
  (curved arrow: fail $\rightarrow$ diagnostic); only queries that pass
  all three stages are executed. The TTQL artifact is produced once
  per graph and reused across all questions.}
  \label{fig:architecture}
\end{figure}

QRAKEN separates graph understanding from question answering into two
phases, sketched in Fig.~\ref{fig:architecture}.

In the \textbf{offline phase} (Fig.~\ref{fig:architecture}, left), the
Graph Distiller explores a SPARQL endpoint and produces a TTQL file
together with a co-occurrence matrix. The TTQL file is not a dump of
triples and not a transcription of the OWL schema: it is a compact
description of the graph as actually populated, containing important
classes, recurring multi-hop paths, conditional pattern frequencies, and
example literals (Sects.~\ref{sec:distill}--\ref{sec:ttql}). The
co-occurrence matrix is kept as a separate machine-side artifact and is
not injected into the LLM prompt.

In the \textbf{online phase} (Fig.~\ref{fig:architecture}, right), the
NTQ (Natural-language To Query) module receives a user question together
with the TTQL file. It prompts an LLM to generate a candidate SPARQL
query, then applies three deterministic correctness checks ---
syntactic, lexical, and data-model --- in the sense of
Sect.~\ref{sec:correctness}. When a check fails, the diagnostic message
is turned into targeted feedback and fed back into the model in a
refinement prompt. The loop iterates until the query passes all three
checks or the iteration budget is exhausted; the surviving query is
then executed against the endpoint and its results returned to the user
(Sect.~\ref{sec:ntq}).

This division of labour invites a fair objection: if the distiller already
queries the graph exhaustively offline, what does the online phase contribute
beyond translating the question and executing the result? Three things, and
deliberately no more. Distillation is question-independent by construction ---
it describes the graph, not the answers --- and the space of admissible
questions is unbounded, so the mapping from a question to a projection, filter
set, aggregation and ordering cannot be precomputed. The correctness checks are
functions of a \emph{candidate query} and cannot exist before one is produced,
even though all the evidence they consult was computed offline. And the online
phase is as thin as we could make it: 1.00--1.43 LLM calls and exactly one
endpoint call per question (Table~\ref{tab:cost}), against the nine or more
model calls of agentic designs. That it reduces to little more than
``translate, verify, execute'' is the intended consequence of moving graph
understanding offline, not an argument against retaining it.

\subsection{Graph distillation as empirical graph summarization}
\label{sec:distill}

QRAKEN's distillation belongs to the broader area of knowledge graph
summarization~\cite{cebiric2019,wang2024} and, more specifically, to
work on linked-data profiling such as ABSTAT~\cite{abstat}, which
abstracts RDF datasets into type- and property-level statistics. TTQL
differs from this lineage in its downstream purpose: it summarizes the
graph simultaneously for \textbf{LLM prompting} and \textbf{deterministic
query validation}, and preserves multi-hop path conditionality and
path-conditioned literal examples that ABSTAT-style one-hop profiles
drop.

\paragraph{The procedure.}
Given a SPARQL endpoint $E$, a maximum hop depth $H$, and a per-class
sample size $N$, the distiller proceeds in five steps. First, it
enumerates the classes present in $E$ and builds a global
\textbf{co-occurrence matrix} $M$ over properties, target classes, and
datatypes by \emph{exhaustive} chunked SPARQL enumeration (not
sampling), so that $M[c,p,c']=0$ is a deterministic certificate of
non-existence of the edge $c\xrightarrow{p}c'$ in~$E$. Second, it ranks
classes by a \textbf{star-shaped score} aggregating instance count,
property diversity, and neighbouring-class diversity via a Borda-style
combination $\mathrm{score}(c) = w_{\mathrm{i}}/r_{\mathrm{inst}}(c) +
w_{\mathrm{p}}/r_{\mathrm{props}}(c) + w_{\mathrm{n}}/r_{\mathrm{nbr}}(c)$
(defaults $0.8/2.5/2.2$), where $r_{\bullet}(c)$ is the rank of $c$ on
each signal and the reciprocal-rank form downweights classes that are
numerous but structurally trivial.
Third, for every ranked class $c$ it draws a stratified sample of $N$
instances that preserves rare properties, and harvests all $n$-hop
patterns up to depth $H$ from those seeds, annotating each pattern
with conditional frequencies and literal examples. Fourth, it assesses
coverage of $M$ against $P$ and issues \textbf{targeted completion}
queries to $E$ for every $(c,\mathrm{edge})$ with coverage $<1$, so
that the final TTQL describes every class-level edge in $M$.
Fifth, it serializes $P$ as a TTQL file
(Sect.~\ref{sec:ttql}, exemplified in Listing~\ref{lst:ttql}) while
keeping $M$ as a separate, machine-side artifact reused later by NTQ
as a deterministic source of truth for data-model checks
(Sect.~\ref{sec:ntq}).

\paragraph{Conditional-frequency semantics.}
Within a pattern, frequencies are interpreted conditionally, in the
spirit of a Markov chain: at the root a frequency answers \emph{``how
often does this class use this property?''}; at nested levels it is
the probability of a sub-pattern \emph{given} that the parent path was
already followed. This conditional form is particularly useful for
join planning, because it exposes the multi-hop combinations that are
actually populated rather than merely possible.

\begin{figure}[t]
\begin{lstlisting}[caption={Excerpt of a TTQL pattern block distilled
from CK25 (\texttt{prod:Hardware}); ``\texttt{...}'' marks elided
siblings or truncated literal lists. Each line starts with a
conditional frequency; \texttt{@literal} leaves carry example values
observed along that path. The two branches shown (via
\texttt{hasCategory} and via \texttt{hasProductManager}, the latter
nested three hops deep) illustrate the path-conditioned structure
TTQL preserves. The full file is in the supplemental material.},label={lst:ttql}]
@prefix prod: <http://ld.company.org/prod-vocab/>

pattern prod:Hardware {
    ?instance-count 100
    ?sample-size    100
    (0.81) prod:compatibleProduct {
        (1.00) prod:Hardware {
            ...
            (0.71) prod:hasCategory {
                (1.00) prod:ProductCategory {
                    (0.35) prod:name { @literal [["Encoder", "Potentiometer", "Transducer", ...]] }
            } }
            (0.71) prod:hasProductManager {
                (0.83) prod:Employee {
                    (0.67) prod:name  { @literal [["Gretel Roth", "Lili Geier", "Ulrik Denzel", ...]] }
                    (0.67) prod:email { @literal [["Gretel.Roth@company.org", "Lili.Geier@company.org", ...]] }
                    (0.67) prod:memberOf {
                        (1.00) prod:Department {
                            (1.00) prod:name { @literal [["Procurement", "Production", "Marketing", ...]] }
                    } } ... }
            ...
    } } } }
\end{lstlisting}
\end{figure}

\subsection{TTQL: a descriptive language for populated graph patterns}
\label{sec:ttql}

TTQL is parsable and human-readable but, unlike Turtle, does not
serialize instance data; it describes the patterns and boundaries of
the graph. Three design choices distinguish it from shape languages.
First, TTQL is \textbf{descriptive rather than prescriptive}: SHACL
and ShEx specify which structures are allowed and whether a graph
matches~\cite{w3c2017shacl,baker2019shex}, and the same prescriptive
stance underlies ontology path patterns derived from CIDOC-CRM
schemas~\cite{mountantonakis2025}, whereas TTQL reports what
is empirically observed, including long-tail or legacy patterns ---
crucial in specialized KGs where the effective data model diverges
from the intended schema.
Second, TTQL is \textbf{natively multi-hop}: shape languages are
typically node-focused, whereas TTQL encodes question-relevant path
patterns (including intermediate joins invisible to a casual reader)
as they occur in data.
Third, TTQL records \textbf{path-conditioned literal examples}: for
every property path, literals actually observed along it, indexed by
root class. This exposes two facets the ontology leaves implicit:
the graph's \emph{empirical lexical convention} (whether a country
surfaces as \texttt{"ITA"}, \texttt{"Italy"}, or \texttt{wd:Q38},
decisive for \texttt{FILTER}s that otherwise silently yield empty
results), and the fact that the same terminal property behaves
differently across roots --- an \texttt{appellation} reached from
\texttt{Artwork}, \texttt{Actor}, or \texttt{Place} carries work
titles, personal names, or toponyms, all collapsed onto
\texttt{xsd:string} by \texttt{rdfs:range} but kept apart by TTQL
coherently with its path-conditional semantics
(Sect.~\ref{sec:distill}). In addition, TTQL is paired with a
\textbf{co-occurrence matrix} reserved machine-side for validation
and never injected into the prompt, so context is not spent on
information that is more useful for deterministic checking than for
generation.

TTQL is therefore not a substitute for SHACL or ShEx but addresses a
different problem. Shapes are node-centred, can be authored from a
schema without ever consulting the data, and govern whether an instance
conforms to a specification; TTQL is always derived from the populated
endpoint, and lets an LLM see how the graph is actually structured
while supplying a compact substrate for data-model checks.

\subsection{NTQ and correctness-guided self-healing}
\label{sec:ntq}

The NTQ module translates natural language into SPARQL using TTQL as
context. The prompt includes the user question, a short explanation of
TTQL syntax, and the relevant TTQL content. The LLM is instructed to
follow the observed property directions and namespaces exactly.

The output then goes through three sequential correctness stages.

\paragraph{Stage 1: Syntactic correctness.}
The candidate query is parsed with \texttt{rdflib}. Before parsing, a
\texttt{PrefixManager} automatically fixes missing or inconsistent
prefix declarations whenever the canonical namespace is known from
TTQL. This removes a frequent but mechanical failure mode.

\paragraph{Stage 2: Lexical correctness.}
All classes and properties used by the query are checked against the
empirical vocabulary extracted from TTQL. This prevents the model from
inventing vocabulary items or using terms outside the observed graph.

\paragraph{Stage 3: Constraint correctness.}
The most important stage. Candidate (class, property, target) triples
in the query are checked against the \textbf{co-occurrence matrix}
distilled from the populated graph (Sect.~\ref{sec:distill}). Because
$M$ is built by exhaustive enumeration rather than sampling,
$M[c,p,c']=0$ is not a low-support heuristic but a \emph{proof of
class-level graph-impossibility}: any triple pattern requiring such an
edge is guaranteed to return an empty binding on the live endpoint,
regardless of instance-level filters. The validator therefore rejects
\textbf{certain-failure} queries before execution --- even when each
individual term is legal at the lexical layer --- and the pipeline is
\textbf{ontology-agnostic} because the ground truth it consults is the
populated data itself, not a declared ontology or shape.

When a stage fails, the diagnostic is turned into targeted feedback
and sent back to the LLM in a refinement prompt. The loop is
therefore \textbf{self-healing}, but unlike purely agentic
designs~\cite{yao2023react} the model is not asked to introspect
freely about its own reasoning: it is asked to revise the query in
light of externally grounded, deterministic diagnostics, which keeps
the repair loop narrow and avoids the context drift that accumulates
when execution traces are fed back unfiltered.

\section{Evaluation}
\label{sec:eval}

\subsection{Text2SPARQL and CK25}

We evaluate QRAKEN on CK25, the corporate benchmark of the First
International Text2SPARQL Challenge: 50 natural-language questions
over a curated KG of 13 classes, 30 properties, $\sim$26.5K
statements, and $\sim$2.6K resources. CK25 is relevant to QRAKEN for
two reasons. First, it is outside the model's pretraining
familiarity, unlike public KGs where memorization blurs
reasoning~\cite{dammu2025} (DBpedia covers 768 classes, Wikidata far
more~\cite{hogan2021});
CK25 remains semantically unfamiliar to the model yet is
\emph{structurally dense} --- its 13 classes and 30 properties
exhibit a bounded but expressive set of recurring multi-hop patterns
that can be faithfully captured by TTQL distillation. Second, the
difficulty on CK25 lies precisely in the composition of those
patterns --- filters, joins, aggregations, and entity-answer
conventions --- rather than in volume, matching QRAKEN's intended
niche of expert-curated KGs with bounded vocabularies and nuanced
query patterns.

\subsection{Experimental protocol}

For the headline run, the CK25 graph was loaded into a local QLever
endpoint (v.\ 1.5.0). The TTQL file was distilled from that endpoint
with per-class sample size $N=100$ and maximum hop depth $H=5$; the
resulting file contains 2{,}181 lines / 122\,KB, which amounts to
$\sim$12\% of the size of the Turtle-serialized graph (970\,KB,
$\sim$26.9K triples) --- i.e.\ the distillation acts as an
$\sim$8$\times$ LLM-oriented compression of the populated data.
The online NTQ stage used \texttt{gpt-4.1-mini} and \texttt{gpt-5.4}
(accessed through the OpenAI API, snapshot dates recorded in each
run's \texttt{meta.json}), data-model validation, up to 5 refinement
iterations, and \texttt{temperature}$=0.1$. Each row is the mean of six
replicates for \texttt{gpt-4.1-mini} and three for every other model
(Sect.~\ref{sec:openweight} covers the open-weight setup).
The reference CK25 gold file and question set are those distributed
by the Text2SPARQL 2025 organizers.

We report two metrics. The main metric is the official \textbf{strict set-based F1} used by
the challenge. The published CK25 gold file does not activate
order-sensitive scoring in a way that changes the aggregate, so for
this dataset the official combined \texttt{F1+NDCG} collapses to the
same set-based F1 in practice. Alongside it we report a \textbf{lenient
F1} used only as a diagnostic. The lenient metric canonicalizes
URI-label equivalences for entity answers; it is informative, but not
directly comparable to the official challenge results.

To compare QRAKEN with the challenge systems under matched conditions,
we re-executed the published \texttt{ck25\_answers.json}
artifacts~\cite{text2sparql2025results} against the same local endpoint
and rescored them with our own evaluator, which (i) runs all candidates
on a single QLever snapshot with identical timeouts, (ii) audits and
deduplicates the published answer files, and (iii) adds a lenient F1
that canonicalises URI--label equivalences where the official gold form
is arbitrary. Relative team ordering is preserved; the comparison below
is therefore a \textbf{matched-condition recomputation on a common local
graph snapshot}, not a re-ranking of the official leaderboard.

\subsection{Results}

\begin{figure}[t]
  \centering
  \includegraphics[width=\textwidth]{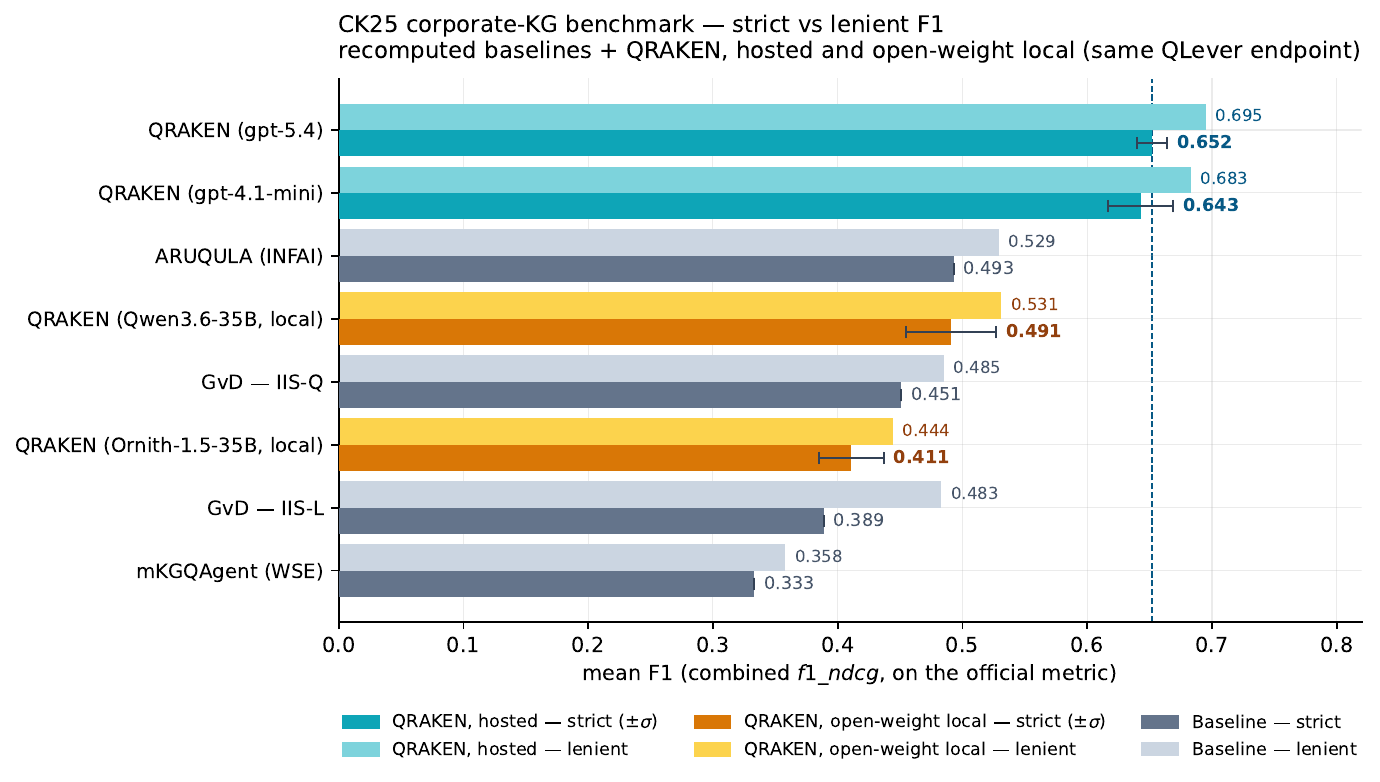}
  \caption{CK25 strict vs lenient F1, matched-condition recomputation on
  a shared QLever endpoint --- the four strongest recomputed participant
  baselines (ARUQULA, IIS-Q, IIS-L, mKGQAgent), the two hosted QRAKEN
  configurations (\texttt{gpt-4.1-mini}, \texttt{gpt-5.4}) and the two
  open-weight QRAKEN configurations run locally at zero marginal cost
  (\texttt{Qwen3.6}, \texttt{Ornith-1.5}), drawn in a separate colour because
  they are not comparable in scale to the hosted models
  (Sect.~\ref{sec:openweight}). Lower-ranked teams (MIPT/AIRI, LACODAM, LABIC,
  FRANZ, AIFB) are omitted for readability and remain in the supplemental
  material. Error bars on QRAKEN rows are 1$\sigma$ across replicates
  ($n{=}6$ for \texttt{gpt-4.1-mini}, $n{=}3$ otherwise).}
  \label{fig:team-comparison}
\end{figure}

Fig.~\ref{fig:team-comparison} reports matched-condition strict and
lenient F1 for the QRAKEN configurations and the four strongest
recomputed participant baselines; QRAKEN bars are the mean of
independent runs at \texttt{temperature=0.1} (six replicates for
\texttt{gpt-4.1-mini}, three otherwise), with 1$\sigma$
error bars across replicates. The two open-weight configurations are
discussed separately in Sect.~\ref{sec:openweight} and are drawn in a
different colour, since they are not comparable in scale to the hosted
models.

With \texttt{gpt-4.1-mini}, QRAKEN outperforms the best matched
recomputed baseline (ARUQULA, same base model family) by \textbf{30\%}
on strict F1; with \texttt{gpt-5.4} the margin is \textbf{32\%}. On
the questions at the \texttt{gpt-4.1-mini} headline,
QRAKEN produces a majority of perfect or partially correct answers,
with the residual zero-score cluster dominated by
\textbf{multi-constraint reasoning} (superlatives combined with
aggregations and multiple filters). Switching to \texttt{gpt-5.4}
closes only a marginal share of that gap ($+0.01$ on aggregate strict
F1) but halves the noise floor ($\sigma = 0.012$ vs $0.026$), which
matters operationally but is not a headline quality driver: the two
configurations are essentially equivalent at the level of mean F1,
and practitioners can choose the cheaper mini model without
significant accuracy loss.

The closest comparator is ARUQULA (both target unknown graphs with
\texttt{gpt-4.1-mini}), and the difference is architectural: ARUQULA
invests in online graph exploration, whereas QRAKEN invests in offline
graph distillation plus online deterministic validation. On a bounded,
curated graph the second trade-off is highly competitive. Similarly,
QRAKEN differs from AIFB's shape-informed prompting not in the
principle of context injection but in its \emph{kind}: shape languages
constrain what should hold, whereas TTQL describes what does hold, how
often, and with which literal values --- empirical content that
matters on a small but densely patterned benchmark.

\subsection{Ablation: what drives the result?}
\label{sec:ablation}

We ran a 2$\times$2 ablation (\texttt{temperature=0.1} on both
models; six replicates per cell on the weaker model and three on the
stronger one) crossing \textbf{prompt context} --- full TTQL patterns
vs.\ the co-occurrence matrix only ($\sim$11$\times$ more compact,
approximating a SHACL-derived shape summary) --- with
\textbf{self-healing}, on or off. We also include a fifth cell (A8)
that replaces the prompt context with an auto-derived SHACL shapes
file (produced from the same CK25 endpoint with
\texttt{shexer}~\cite{aifb2025} and filtered to the \texttt{prod:}
namespace), giving a direct comparison against the shape-informed
baseline of the AIFB team. The validator accepts $\sim$94\% of
first-iteration queries; on the remaining $\sim$6\% the loop either
refines the query to a graph-grounded form or exhausts its iteration
budget.

\begin{table}[t]
\caption{Ablation on CK25 (strict F1, mean $\pm$ sd). Self-healing is on with a
budget of 5 iterations and off at 1. The first two model columns are the hosted
models of the headline runs; the last two are open-weight 35B MoE models
quantised to 4 bits and run locally (Sect.~\ref{sec:openweight}). Row A8 reports
an auto-derived SHACL baseline as a direct comparator for shape-informed
prompting~\cite{aifb2025}; it was not run on \texttt{gpt-5.4}. The three summary
rows give the marginal contribution of the TTQL axis (A2 $-$ A4), of the
self-healing loop (A1 $-$ A2), and of TTQL over the shape baseline (A2 $-$ A8).}
\label{tab:ablation}
\centering
\scriptsize
\renewcommand{\arraystretch}{1.15}
\setlength{\tabcolsep}{3pt}
\begin{tabular}{l l c r r r r}
\toprule
\textbf{Cell} & \textbf{Prompt context} & \textbf{Heal.} &
\textbf{gpt-4.1-mini} & \textbf{gpt-5.4} &
\textbf{Ornith-1.5} & \textbf{Qwen3.6} \\
 & & & ($n{=}6$) & ($n{=}3$) & ($n{=}3$) & ($n{=}3$) \\
\midrule
A1 & Full TTQL patterns        & on  & \textbf{0.643} $\pm$ \textbf{0.026} & \textbf{0.652} $\pm$ \textbf{0.012} & 0.411 $\pm$ 0.026 & \textbf{0.491} $\pm$ \textbf{0.036} \\
A2 & Full TTQL patterns        & off & 0.617 $\pm$ 0.013 & 0.654 $\pm$ 0.010 & 0.420 $\pm$ 0.008 & 0.424 $\pm$ 0.028 \\
A3 & Co-occ.\ matrix only      & on  & 0.359 $\pm$ 0.016 & 0.490 $\pm$ 0.032 & 0.327 $\pm$ 0.020 & 0.351 $\pm$ 0.016 \\
A4 & Co-occ.\ matrix only      & off & 0.308 $\pm$ 0.015 & 0.501 $\pm$ 0.012 & 0.284 $\pm$ 0.017 & 0.309 $\pm$ 0.005 \\
A8 & SHACL-auto (shexer)       & off & 0.376 $\pm$ 0.018 & --- & 0.255 $\pm$ 0.012 & 0.203 $\pm$ 0.011 \\
\midrule
\multicolumn{3}{l}{\textbf{TTQL contribution} (A2 $-$ A4)}
   & \textbf{+0.309} & \textbf{+0.153} & \textbf{+0.136} & \textbf{+0.115} \\
\multicolumn{3}{l}{\textbf{Self-healing contribution} (A1 $-$ A2)}
   & \textbf{+0.026} & \textbf{$-$0.003} & \textbf{$-$0.009} & \textbf{+0.067} \\
\multicolumn{3}{l}{\textbf{TTQL vs SHACL-auto} (A2 $-$ A8)}
   & \textbf{+0.241} & --- & \textbf{+0.165} & \textbf{+0.221} \\
\bottomrule
\end{tabular}
\end{table}

The ablation isolates three separable effects.

\emph{TTQL pattern blocks are the dominant and robust driver of
performance.} Replacing the full TTQL file with its co-occurrence
matrix alone ($\sim$11$\times$ more compact, approximating a
SHACL-derived shape summary) costs \textbf{+0.309} strict F1 on
\texttt{gpt-4.1-mini} and \textbf{+0.153} on \texttt{gpt-5.4}; the gap
is an order of magnitude larger than any within-condition noise on
either model and is robust to sample size ($n{=}6$ per cell on the
weaker model). The weaker the model's prior, the more it benefits
from the empirical pattern information: gpt-4.1-mini \emph{doubles}
its F1 when moving from a shape-only to a pattern-level prompt
--- the value of TTQL is descriptive content, not compactness.

\emph{Pattern-level content outperforms shape-level content.} The
direct comparator A8 --- an auto-derived SHACL shapes file with the
same iteration budget as A2 --- lands at $0.376\pm 0.018$, only
$+0.07$ above the co-occurrence-matrix-only baseline A4 and still
$\mathbf{-0.241}$ absolute ($\mathbf{+64\%}$ relative in A2's favor)
below TTQL patterns at identical budget: giving the LLM a positive,
empirically-grounded description of how the graph is actually used
yields roughly $4.5\times$ the quality lift of exposing the same schema
as structural constraints. This
directly addresses the AIFB family of shape-informed
baselines~\cite{aifb2025}: on CK25 the kind of graph context
injected matters far more than the mere presence of graph-aware
context.

\emph{Self-healing is a certain-failure filter, not a generic
optimiser.} Because the co-occurrence matrix is exhaustive
(Sect.~\ref{sec:distill}), Stage~3 rejects only queries with a
zero-support triple pattern: the loop intervenes exactly when the
first-shot query is provably empty against the target graph, and does
nothing otherwise. Across the four models its contribution ranges from
$-0.009$ to $+0.067$ (A1 $-$ A2) at a per-question overhead below
\$0.01 (Table~\ref{tab:cost}) --- bounded upside \textbf{by design}
where first-shot quality is already high, and no systematic regression
where it is not. Its floor is ``no cost'', its ceiling the elimination
of guaranteed-empty answers, rather than a general-purpose quality
boost.

\subsection{Open-weight models}
\label{sec:openweight}

To test whether the approach survives outside frontier hosted models, we re-ran
the full ablation grid with two open-weight 35B-A3B mixture-of-experts models
quantised to 4 bits (\texttt{Ornith-1.5} and \texttt{Qwen3.6}), served locally
through an MLX runtime on a single laptop: 30 runs, 1{,}500 questions,
5\,h\,43\,m of wall clock, and \textbf{zero marginal cost}. The pipeline is
unchanged and remains training-free; only the model behind the NTQ module
differs.

Both TTQL claims hold on open weights (Table~\ref{tab:ablation}). The TTQL axis
contributes $+0.136$ and $+0.115$ strict F1 (A2 $-$ A4), in line with the
$+0.153$ of \texttt{gpt-5.4}, and TTQL stays ahead of the auto-derived SHACL
baseline by $+0.165$ and $+0.221$ (A2 $-$ A8), the latter close to the $+0.241$
measured on \texttt{gpt-4.1-mini}. The effect is therefore attributable to the
\emph{kind} of context injected rather than to the capability of a particular
model. In absolute terms \texttt{Qwen3.6} reaches $0.491\pm0.036$ strict F1, on
par with the strongest recomputed participant system (ARUQULA, $0.493$) and
ahead of the other eight, on consumer hardware and at no per-question cost
(Fig.~\ref{fig:team-comparison}). The residual gap to the hosted configurations
is not schema comprehension but SPARQL validity and answer shape: structurally
sound queries that filter on a literal the graph does not contain, and entity
URIs projected where the gold answer holds a literal. Both are pipeline-side
losses rather than a model-capability ceiling.

\subsection{Cost analysis}
\label{sec:cost-analysis}

Because Sect.~\ref{sec:cost} criticizes other Text2SPARQL systems for
not reporting cost, we do the same accounting for QRAKEN on the same
50-question CK25 benchmark, restricted to the \emph{online} per-query
path (the upstream distillation is a one-off offline step, $\sim$15\,min
of wall-clock for the whole graph, amortized across all questions).
Endpoint calls are $1$ per question: the validator consults the
pre-distilled matrix machine-side and never round-trips to SPARQL,
and the validated query is executed once by the runner. The costed
dimension is therefore LLM calls $\times$ tokens at public OpenAI rates
(\$0.40/\$1.60 per $10^6$ for \texttt{gpt-4.1-mini}, a conservative
\$2.00/\$8.00 for \texttt{gpt-5.4}), plus median wall-clock latency per
question (median rather than mean to guard against client-side stalls).
Table~\ref{tab:cost} reports the QRAKEN configurations and the
external systems.

\begin{table}[t]
\caption{Online cost per question on CK25, aggregated over the same
runs that back Table~\ref{tab:ablation}. \textbf{Calls/q} is the mean
number of LLM calls per question (self-healing iterations);
\textbf{tokens/q} is the mean total of input+output tokens;
\textbf{\$/100q} multiplies total per-question cost by 100;
\textbf{lat./q} is median wall-clock per question. The open-weight row is the
local grid of Sect.~\ref{sec:openweight}, where per-call accounting was not
instrumented and latency is warm-cache. For ARUQULA, calls/q is the mean number
of \emph{agent steps} and lat./q the \emph{mean} time to answer on the corporate
dataset, both as reported in its own evaluation~\cite{aruqula2025}.}
\label{tab:cost}
\centering
\scriptsize
\renewcommand{\arraystretch}{1.15}
\setlength{\tabcolsep}{4pt}
\begin{tabular}{l l r r r r}

\toprule
\textbf{Cell} & \textbf{Model} &
\textbf{Calls/q} & \textbf{Tokens/q} &
\textbf{\$/100q} & \textbf{Lat./q} \\
\midrule
A1 (full, healing on)  & gpt-4.1-mini & 1.43 & 41{,}905 & \$1.69 & 3.5\,s \\
A2 (full, healing off) & gpt-4.1-mini & 1.00 & 29{,}164 & \$1.18 & 5.0\,s \\
A1 (full, healing on)  & gpt-5.4      & 1.21 & 34{,}933 & \$7.08 & 3.2\,s \\
A2 (full, healing off) & gpt-5.4      & 1.00 & 29{,}182 & \$5.89 & 2.5\,s \\
\midrule
A3 (shape only, healing on)  & gpt-4.1-mini & 1.40 & 6{,}757 & \$0.29 & 2.6\,s \\
A4 (shape only, healing off) & gpt-4.1-mini & 1.00 & 4{,}711 & \$0.20 & 2.6\,s \\
\midrule
A1 (full, healing on) & open-weight, local & n/a & $\sim$29{,}000 & \$0.00 & 8--11\,s \\
\midrule
\multicolumn{2}{l}{\emph{External, as reported by competitors:}} & & & & \\
\multicolumn{2}{l}{ARUQULA (INFAI, gpt-4.1-mini)~\cite{aruqula2025}}
                                              &  9.9 &  n/a & n/a             & 59.6\,s \\
\multicolumn{2}{l}{mKGQAgent (GPT-3.5 / GPT-4o)~\cite{perevalov2025}}
                                              & 13.0 &  n/a & \$0.48 / \$3.06 & n/a \\
\multicolumn{2}{l}{SPARQL-LLM (default model)~\cite{sparqlllm2025}}
                                              &  n/a &  n/a & $\sim$\$1.00     & n/a \\
\bottomrule
\end{tabular}
\end{table}

QRAKEN's headline A2 at \$1.18/100q is comparable to SPARQL-LLM
($\sim$\$1/100q) and $2.6\times$ cheaper than mKGQAgent/GPT-4o
(\$3.06/100q) while reaching substantially higher strict F1 (0.617 vs
0.333, Fig.~\ref{fig:team-comparison}), because the distiller
amortizes graph understanding offline. Self-healing adds
$\sim$\$0.5/100q for $+0.026$ F1; full TTQL injects $5\times$ more
tokens yet stays under \$0.02/question and $5$\,s latency, so the
$+0.31$ TTQL contribution is a quality gain, not a quality-for-price
trade-off.

The sharpest comparison is with ARUQULA, which runs the same base model on the
same benchmark. On the corporate dataset it reports a mean of $9.92$ agent
steps and $59.6$\,s per question~\cite{aruqula2025}; every step passes through
a controller LLM call, and its published action counts amount to roughly nine
backend calls per question, about three of which execute SPARQL. QRAKEN's A1
answers the same questions with $1.43$ LLM calls and a \emph{single} endpoint
call at $3.5$\,s median latency, at $0.643$ versus $0.493$ strict F1. ARUQULA
reports neither tokens nor monetary cost, so the comparison is confined to the
dimensions it does report; the two profiles are structurally different, since
QRAKEN pays once for a long but cache-friendly TTQL prefix whereas online
exploration pays repeatedly for a growing reasoning trace. On open weights the
monetary axis disappears altogether: the same configuration costs nothing per
question and trades latency instead ($\sim$8--11\,s warm).

\subsection{Error analysis}
\label{sec:erroranalysis}

On the evaluable CK25 questions at the \texttt{gpt-4.1-mini}
headline, the 16 zero-score answers cluster around three recurring
failure modes identified from the archived audit files
(Sect.~\ref{sec:supplemental}): \textbf{(i) multi-constraint
composition} --- superlatives combined with aggregation and two or
more filters, the largest share, which \texttt{gpt-5.4} substantially
reduces; \textbf{(ii) literal canonicalisation} when several lexical
conventions co-exist for the same property (e.g.\ mixed English/German
labels), where TTQL literal examples are helpful but not always
sufficient; and \textbf{(iii) aggregation-scope errors} (correct
predicates, wrong \texttt{GROUP BY}/\texttt{HAVING}), which the current
validator does not catch; \emph{path shortening}
(Sect.~\ref{sec:correctness}) is a fourth, rarer mode, caught by the
data-model validator in most cases.

\subsection{Threats to validity}
\label{sec:threats}

All quantitative claims rest on a \textbf{single benchmark} (CK25, 50
questions). The open-weight grid broadens the evidence along the \emph{model}
axis --- four configurations instead of two --- but not along the \emph{graph}
axis, and validation on a larger, more heterogeneous graph remains the most
important missing experiment. The \textbf{shape baseline} (A3/A4) approximates a SHACL/ShEx summary by stripping TTQL to the
co-occurrence matrix; a full replica of a shape-prompting
system~\cite{aifb2025} would be a stronger comparator. For the hosted models,
API snapshot dates are recorded in every run artifact; determinism is
guaranteed at the level of generated queries, not model internals.

We also report a \textbf{single distillation setting} ($N{=}100$, $H{=}5$) and
do not measure F1 as a function of those parameters. The deterministic half of
the pipeline is insensitive to them by construction: $M$ is built by exhaustive
enumeration rather than sampling, so the Stage-3 impossibility certificate does
not depend on $N$ or $H$, and the targeted-completion step
(Sect.~\ref{sec:distill}) forces the emitted TTQL to describe every class-level
edge of $M$ whatever the sample size. The two parameters therefore govern only
the richness of pattern blocks and literal examples --- the generative context
--- but how much that richness is worth is an empirical question we have not
answered here, and a sensitivity sweep over $N$ and $H$ is the first experiment
on our list.

\section{Conclusion, Applicability, and Future Work}
\label{sec:conclusion}

We argued that the core challenge in natural-language interfaces to
knowledge graphs is not query generation per se but correctness under
graph-specific design choices, and that QRAKEN addresses it by
distilling the graph into an empirical LLM-readable summary (TTQL)
and validating generation online through deterministic checks --- a
descriptive alternative to both prescriptive shape languages and
online agentic exploration. On this single, relatively small benchmark (CK25, 50 questions), under
matched-condition recomputation on a locally re-loaded snapshot, QRAKEN
attains the highest strict F1 among the recomputed participant systems
at the same base model family, suggesting that in domain-specific KGQA
graph distillation plus deterministic validation can be more decisive
than heavier online agent orchestration --- a finding that warrants
validation on additional benchmarks.

QRAKEN is strongest on curated, semantically rich KGs with bounded
vocabularies and recurring multi-hop patterns --- cultural-heritage,
scientific, and corporate graphs. The distiller itself is linear in
class--property pairs and graph-size-agnostic; the ceiling is instead
\textbf{monolithic TTQL injection}, which saturates when the distilled
summary of a KG with thousands of heterogeneous classes no longer fits
usefully in a single prompt. The limit is concrete and measurable: CK25's TTQL
is $\sim$30K tokens ($\sim$29.1K once compacted), already close to a third of
the context window of the open-weight models of Sect.~\ref{sec:openweight}.
Its \emph{price}, by contrast, is largely avoidable --- placing the TTQL block
at the head of the prompt makes it KV-cacheable, which we measure at 97\% reuse
locally (151\,s cold against $\sim$11\,s warm per run) and which hosted
providers discount as well, so prompt order is a cost lever on both paths. What
does not scale is the context budget rather than the distillation. The most
promising next steps are
therefore a \textbf{retrieval layer over TTQL} (selecting relevant
fragments per question) and a dedicated
\textbf{multi-constraint reasoning} layer addressing the dominant
CK25 failure mode. Broader evaluation on further private and
scholarly graphs is ongoing.

\paragraph*{Supplemental Material Statement:}\label{sec:supplemental}
A reproducibility package at
\url{https://github.com/RemoGrillo/qraken_ck25} contains
the \textbf{TTQL file}, the \textbf{NTQ module}, the \textbf{LLM gateway},
the \textbf{benchmark engine}, the
\textbf{CK25 dataset} and competitor answers backing
Fig.~\ref{fig:team-comparison}, and raw \textbf{run artifacts} for every
ablation cell with generated SPARQL, endpoint bindings, chart
script, and a SHA-256 manifest. All paths are documented in the package.
The offline \textbf{Graph Distiller} itself is
withheld as a partial snapshot of a broader evolving platform, but its
CK25 output is released in full and Sect.~\ref{sec:distill} supports
re-implementation, so all empirical claims remain independently verifiable.


\section*{Declaration on Generative AI}
During the preparation of this work, the authors used Anthropic Claude
(Sonnet and Opus 5) in order to: Grammar and spelling check, Paraphrase
and reword, Improve writing style, Generate code. After using these
tool(s)/service(s), the authors reviewed and edited the content as needed
and take full responsibility for the publication's content.
Generative AI was not used
to fabricate experimental results, synthesize benchmark data, invent
references, or produce analyses of runs that were not actually executed.
Large language models are studied \emph{as objects} in this paper (they
are part of the pipeline under evaluation); their use inside the QRAKEN
pipeline is fully described in Sect.~\ref{sec:method} and is not covered
by this declaration.

\label{lastbodypage}
\bibliography{references}

\end{document}